\documentclass{article}
\usepackage{preamble}
\title{A Decentralized Reinforcement Learning Framework for Efficient Passage of Emergency Vehicles}

\author{%
  Haoran Su$^{\dagger}$\thanks{This study is conducted with his internship with Siemens. He is also sponsored by DDETFP by U.S. Department of Transportation Federal Highway Administration.} , Yaofeng Desmond Zhong$^{\S}$, Biswadip Dey$^{\S}$, Amit Chakraborty$^{\S}$ \\
  $^{\dagger}$ New York University | $^{\S}$ Siemens Corporation, Technology\\
  \texttt{haoran.su@nyu.edu}, (\texttt{yaofeng.zhong}, \texttt{biswadip.dey}, \texttt{amit.chakraborty})\texttt{@siemens.com}
}

\usepackage{hyperref, lipsum}   

\begin{document}
\setlength{\abovedisplayskip}{1pt}
\setlength{\belowdisplayskip}{1pt}

\maketitle

\begin{abstract}
Emergency vehicles (EMVs) play a critical role in a city's response to time-critical events such as medical emergencies and fire outbreaks. The existing approaches to reduce EMV travel time employ route optimization and traffic signal pre-emption without accounting for the coupling between route these two subproblems. As a result, the planned route often becomes suboptimal. In addition, these approaches also do not focus on minimizing disruption to the overall traffic flow. To address these issues, we introduce \textit{EMVLight} in this paper. This is a decentralized reinforcement learning (RL) framework for simultaneous dynamic routing and traffic signal control. EMVLight extends Dijkstra's algorithm to efficiently update the optimal route for an EMV in real-time as it travels through the traffic network. Consequently, the decentralized RL agents learn network-level cooperative traffic signal phase strategies that reduce EMV travel time and the average travel time of non-EMVs in the network. We have carried out comprehensive experiments with synthetic and real-world maps to demonstrate this benefit. Our results show that EMVLight outperforms benchmark transportation engineering techniques as well as existing RL-based traffic signal control methods.
\end{abstract}

\section{Introduction}
As any reduction in the travel time of an EMV increases the likelihood that it would save someone's life or reduce property loss, there is a significant benefit in reducing the average EMV travel time on increasingly crowded roads. To tackle this challenge, existing approaches rely upon -- \textit{route optimization} which uses graph-theoretic tools to find the fastest path; and \textit{traffic signal pre-emption} which alters the traffic signals to prioritize EMV passage and improves overall safety at the intersections. Please see Section~S1 for additional details about the existing approaches on route optimization and traffic signal control. However, the constantly changing traffic conditions can significantly affect the optimality of a route prescribed by route optimization. In addition, traffic signal pre-emption also changes the optimal route by causing serious disruptions to the traffic flow.

To address these issues and achieve efficient EMV passage, we introduce \textbf{EMVLight} which is a decentralized multi-agent RL framework with a dynamic routing algorithm to control traffic signal phases. We introduce a computationally efficient dynamic routing algorithm and an RL-based traffic signal pre-emption strategy to not only reduce \emph{EMV travel time} but also to reduce the \emph{average travel time of non-EMVs}. Our experiments demonstrate that EMVLight outperforms traditional traffic engineering methods and existing RL methods on both metrics under different traffic configurations.

\section{Application Context}
Emergency response time is the key indicator of a city's emergency management capability. Reducing response time saves lives and prevents property loss. For instance, the survivor rate from a sudden cardiac arrest without treatment drops 7\% - 10\% for every minute elapsed, and there is barely any chance to survive after 8 minutes \cite{nichol2008regional}. The time interval for an \emph{emergency vehicle (EMV)}, e.g., ambulances, fire trucks, and police cars, to travel from a rescue station to an incidence site, i.e., the \textit{EMV travel time}, accounts for a significant portion of the emergency response time. However, overpopulation and
urbanization have been exacerbating road congestion, making it more and more challenging to reduce the average EMV travel time \cite{end-to-end-response-times}. There is a severe urgency and significant benefit in reducing the average EMV travel time on increasingly crowded roads.
%
%
%
%
%
%

%
%

\section{Preliminaries}

\textbf{Traffic map, Link, Lane:}
A traffic map can be represented by a graph $G(\mathcal{V}, \mathcal{E})$, with intersections as
\begin{wrapfigure}[24]{r}{0.41\textwidth}
  \begin{center}
  \vspace{-0.5em}
    \includegraphics[width=0.3\textwidth]{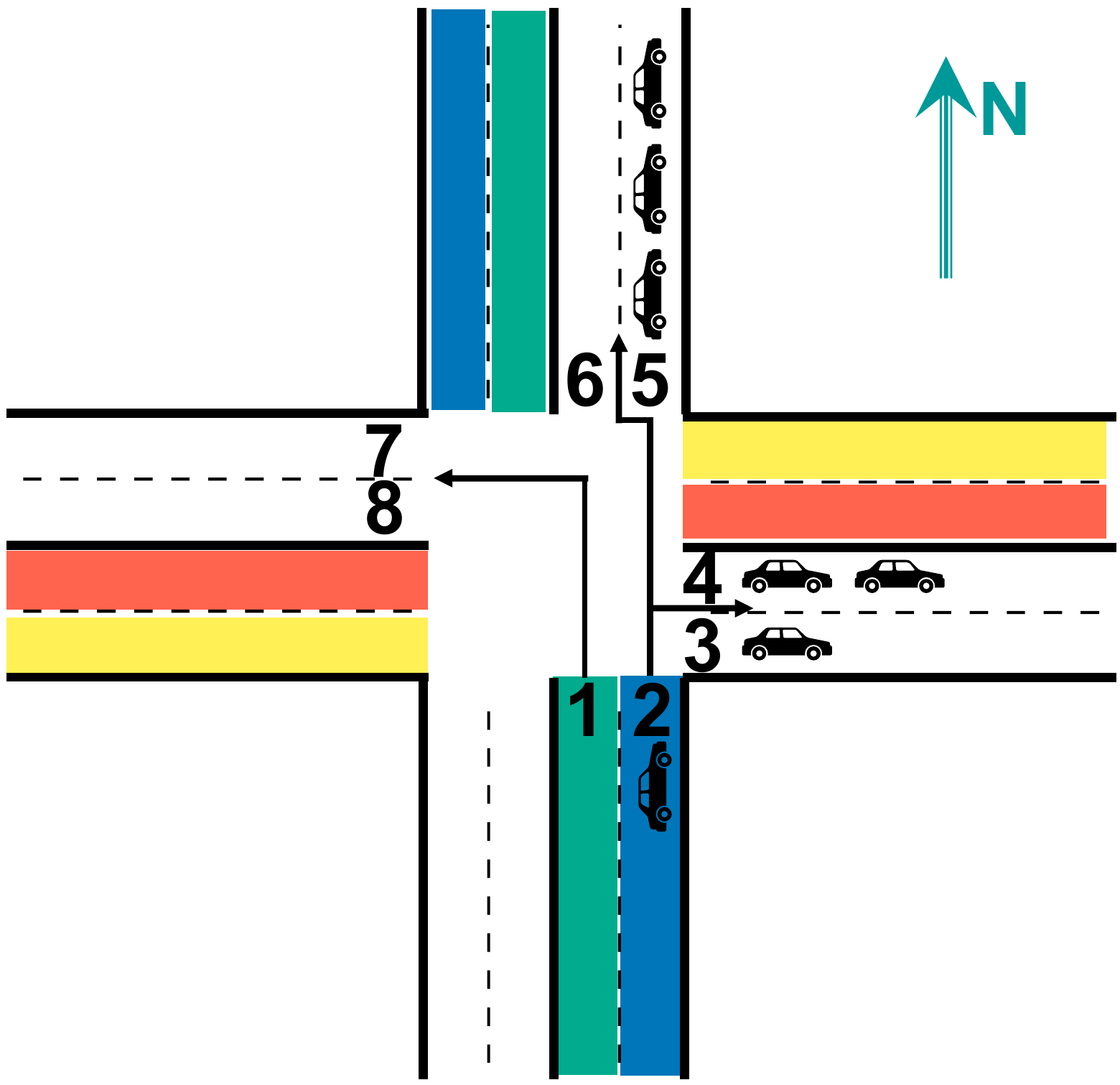}
  \vspace{-1.05em}
  \end{center}
  \caption{\small{This figure illustrates traffic movements through an intersection and an example pressure calculation for incoming lane \#2. Vehicles on Lane 1 are turning left, and vehicles on Lane 2 may either go straight or turn right. After turning right, vehicles can enter either lane. Thus, the incoming link from South has the following traffic movement: {(1, 7), (1, 8), (2, 5), (2, 6),  (2, 3), (2, 4)}. Assuming each lane has the maximum capacity of 5, the lane pressure for Lane 2 can then be calculated as: $w(2) = |1/5 - (1/2)*(1/5 + 2/5) - (1/2)*(3/5 + 0/5)| = 0.4$.}}
  \label{fig_movements}
  \vspace{-1.5em}
\end{wrapfigure}
nodes and road segments between intersections as edges. We refer to a one-directional road segment between two intersections as a link. A link has a fixed number of lanes, denoted as $h(l)$ for lane $l$. Fig.~\ref{fig_movements} shows 8 links and each link has 2 lanes.

\begin{wrapfigure}[14]{r}{0.41\textwidth}
  \begin{center}
  \vspace{1em}
    \includegraphics[width=0.41\textwidth]{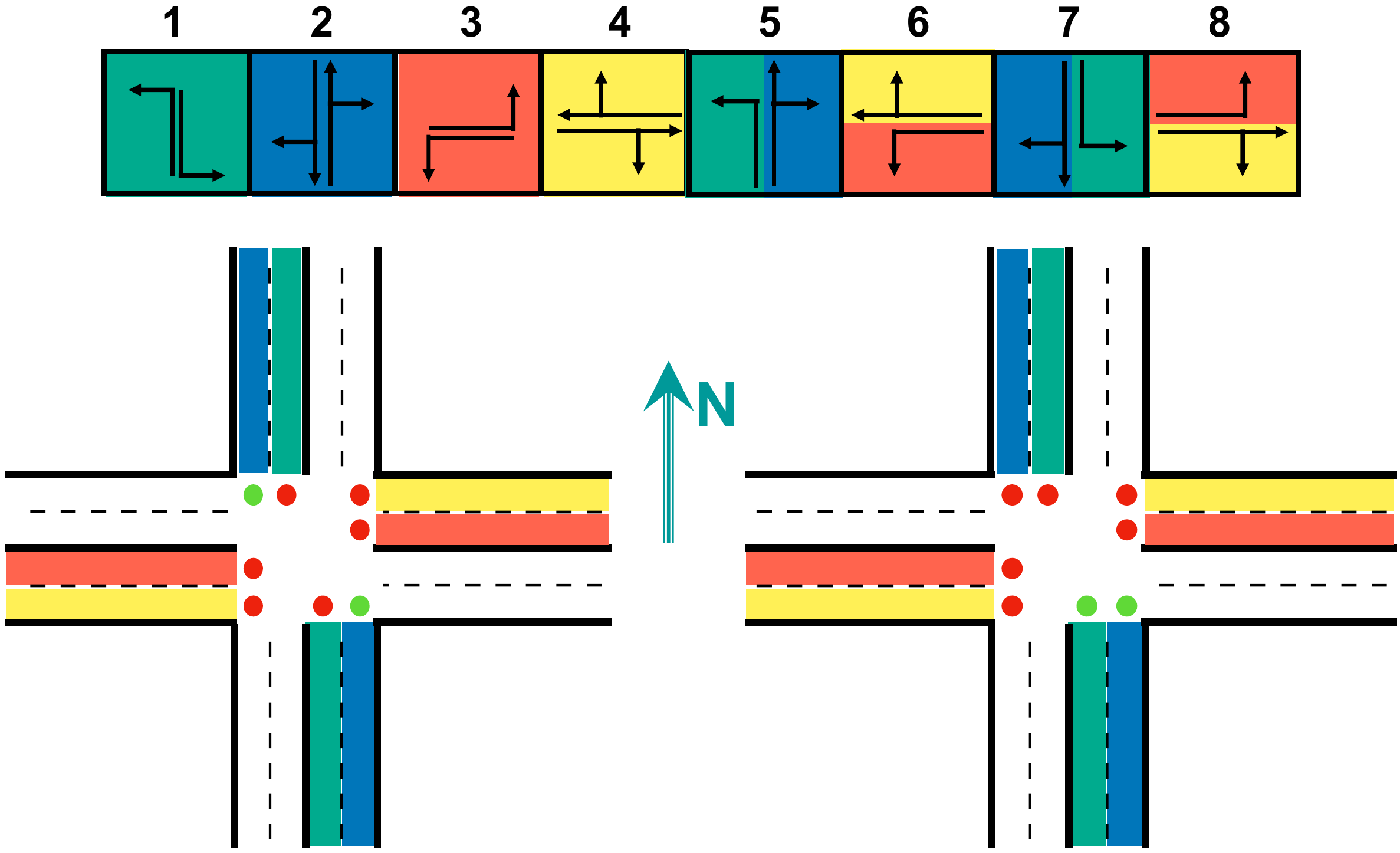}
  \vspace{-1.25em}
  \end{center}
  \caption{\small{\emph{Top} panel of this figure shows the 8 signal phases for a 4-way intersection. The \emph{bottom-left} and \emph{bottom-right} panels show phase 2 and 5, respectively.}} \label{fig_phases}
\end{wrapfigure}

\textbf{Traffic Movements:}
A traffic movement $(l,m)$ is defined as the traffic traveling across an intersection from an incoming lane $l$ to an outgoing lane $m$. The intersection shown in Fig.~\ref{fig_movements} has 24 permissible traffic movements. The set of all permissible traffic movements of an intersection is denoted as $\mathcal{M}$.

\textbf{Traffic Signal Phase:}
A traffic signal phase is defined as the set of permissible traffic movements. As shown in Fig.~\ref{fig_phases}, an intersection with 4 links has 8 phases.

\textbf{Pressure of an incoming lane:}
The pressure of an incoming lane $l$ measures the unevenness of vehicle density between lane $l$ and corresponding outgoing lanes in permissible traffic movements. 
The vehicle density of a lane is $x(l)/x_{max}(l)$, where $x(l)$ is the number of vehicles on lane $l$ and $x_{max}(l)$ is the vehicle capacity on lane $l$, which is related to the length of a lane. Then the pressure 

of an incoming lane $l$ is 
\begin{displaymath} 
w(l) = \left|\frac{x(l)}{x_{max}(l)} - \sum_{\{m|(l, m)\in \mathcal{M}\}}\frac{1}{h(m)}\frac{x(m)}{x_{max}(m)}\right|,
\end{displaymath}
where $h(m)$ is the number of lanes of the outgoing link which contains $m$. This concept has been explained in Fig.~\ref{fig_movements} wherein $h(m)=2$ for all the outgoing lanes.

\textbf{Pressure of an intersection:}
The pressure of an intersection ($P$) is the average of the pressure of all incoming lanes. It indicates the unevenness of vehicle density between incoming and outgoing lanes in an intersection. Intuitively, reducing the pressure leads to more evenly distributed traffic, which indirectly reduce congestion and average travel time of vehicles.
\section{Dynamic Routing}
%
%
%
%
%
A traffic network can be perceived as a weighted graph with the weights representing the \emph{intra-link travel time}, i.e., the EMV travel time along each link which has been estimated from the number
of vehicles on that link. This enables the use of Dijkstra's algorithm for finding solutions to the EMV routing problem. However, continuously changing traffic conditions keep changing the intra-link travel time and thereby render the weighted graph dynamic. Although this challenge can be addressed by running Dijkstra's algorithm repeatedly as the EMV travels through the network, it is not viable from a computational perspective. In this work, we extend Dijkstra's algorithm to efficiently update the optimal route based on the most recent information about the intra-link travel times (Algorithm~\ref{alg:ETA_prepopulation}). We first run Dijkstra's algorithm to obtain the \textit{estimated time of arrival (ETA)} from each intersection to the destination and the \textit{next intersection} along the shortest path. Since a sequence of processes, including call-taker processing, are performed before an EMVs is dispatched, it is reasonable to assume that this process can be done before an EMV starts to travel. Once an EMV has started its journey, we 
\begin{wrapfigure}{r}{0.525\textwidth}
\vspace{-1.0em}
\begin{algorithm}[H]
    \caption{Dynamic Dijkstra's for EMV routing}
    \label{alg:ETA_prepopulation}
    \SetEndCharOfAlgoLine{}
    \SetKwInOut{Input}{Input}
    \SetKwInOut{Output}{Output}
    \SetKwData{ETA}{ETA}
    \SetKwData{Next}{Next}
    \SetKwFor{ParrallelForEach}{foreach}{do (in parallel)}{endfor}
    \Input{\\\hspace{-3.7em}\vspace{-1em}
        \begin{tabular}[t]{l @{\hspace{3.3em}} l}
        $G=(\mathcal{V}, \mathcal{E})$ & traffic map as a graph \\
        $T^t = [T_{ij}^t]$ & intra-link travel time at time $t$ \\
        $i_d$  & index of the destination
        \end{tabular}
    }
    \Output{\\\hspace{-3.7em}\vspace{-1em}
        \begin{tabular}[t]{l @{\hspace{1.5em}} l}
        $\mathsf{ETA}^t = [\mathsf{ETA}^t_i]$ & ETA of each intersection \\
        $\mathsf{Next}^t = [\mathsf{Next}^t_i]$ & next intersection to go \\
        & from each intersection
        \end{tabular}
        
    }
    \vspace{1.0em}
    \tcc{pre-population}
    $\mathsf{ETA}^0, \mathsf{Next}^0$ $=$ \texttt{Dijkstra}$(G, T^0, i_d)$\;
    \vspace{0.5em}
    \tcc{dynamic routing}
    \For{$t = 0 \to T$}{
        \ParrallelForEach{$i \in \mathcal{V}$}{
            $\mathsf{ETA}_i^{t+1} \gets \min_{(i, j)\in \mathcal{E}} (\mathsf{ETA}_j^t + T_{ji}^t)$\;
            $\mathsf{Next}_i^{t+1} \gets \arg\min_{\{j|(i, j)\in \mathcal{E}\}}(\mathsf{ETA}_j^{t} + T_{ji}^t$)
            }
            }
\end{algorithm}
\vspace{-3.0em}
\end{wrapfigure}
update $\mathsf{ETA}$ and $\mathsf{Next}$ values for each intersection. As these updates only depend on information about neighboring intersections, they are computationally inexpensive and can be executed in parallel. Please see Section~S5 for further details about how intra-link travel time is estimated in real time.
\section{Setting up the RL Problem}
While dynamic routing directs the EMV to the destination, it does not take into account the possible wait times due to red lights at the intersections. Thus, traffic signal pre-emption is also required for the EMV to arrive at the destination in the least amount of time. However, since traditional pre-emption only focuses on reducing the EMV travel time, it can significantly increase the average travel time of non-EMVs. In this work, we set up traffic signal control for efficient EMV passage as a decentralized RL problem. In our formulation, we employ multiple RL agents (one for each intersection) that work cooperatively to reduce both EMV travel time and the average travel time of non-EMVs.

\subsection{Types of agents for EMV passage}
\begin{wrapfigure}[11]{r}{0.45\textwidth}
\begin{center}
\vspace{-3em}
    \includegraphics[width=0.375\textwidth]{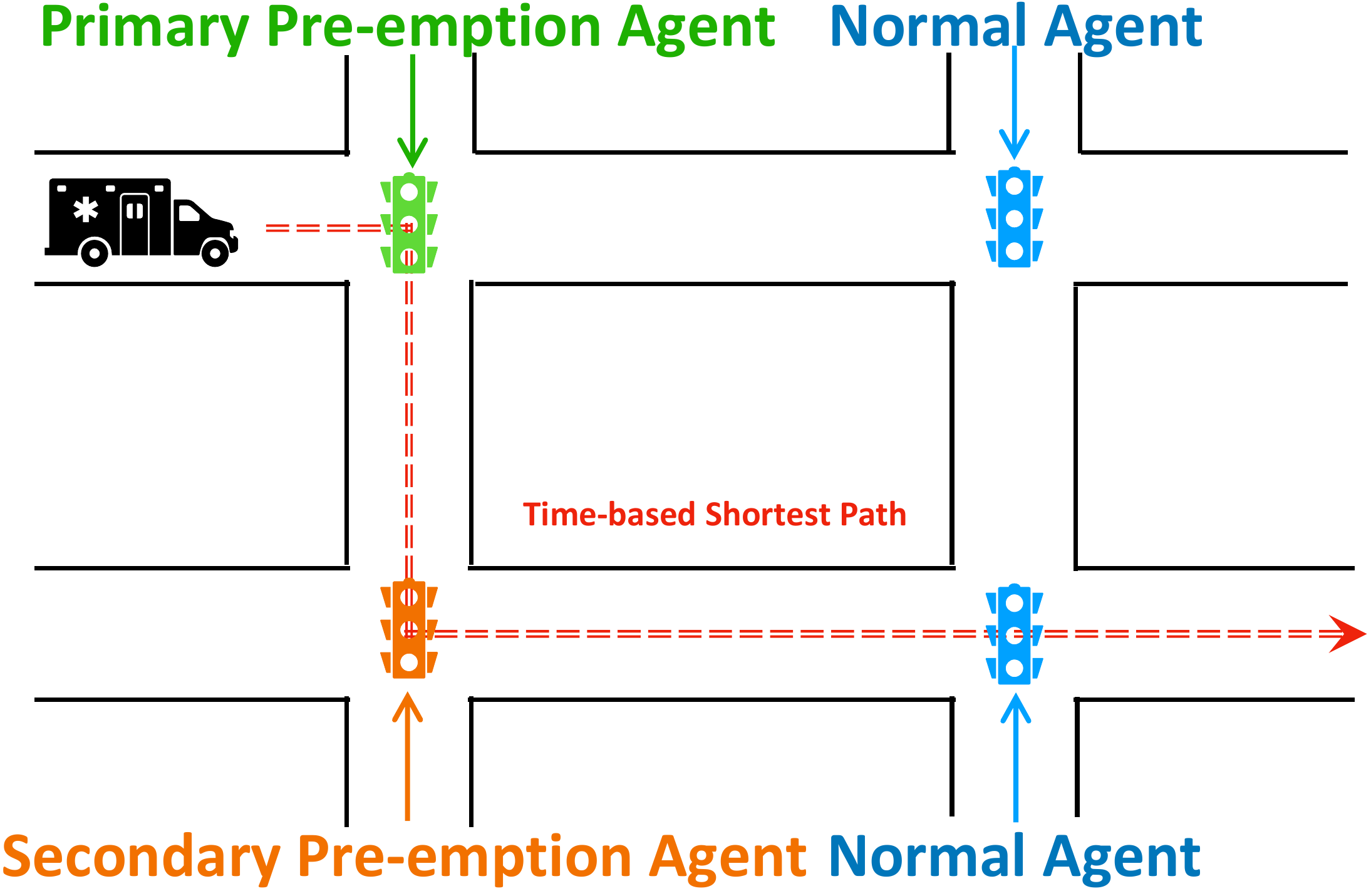}
\vspace{-1.1em}
\end{center}
\caption{\small{EMVLight employs 3 types of agents.}}
\vspace{-1em}
\label{fig_secondary}
\end{wrapfigure}

Whenever an EMV is travelling through the traffic network, the agents that control the traffic signal at
the intersections are grouped into 3 distinct categories based on EMV location and routing (Fig.~\ref{fig_secondary}). An agent is categorized as a \emph{primary pre-emption agent} $i_p$ if an EMV is travelling on one of its incoming links. The agent at the next intersection along the optimal route is categorized as a \emph{secondary pre-emption agent}, i.e. $i_s = \mathsf{Next}_{i_p}$. The rest of the agents are \emph{normal agents}. Each of these agent types has different local goals which is captured into their reward functions.

\subsection{Agent design}
\textbf{$\bullet$ State:} The state of an agent $i$ at time $t$ includes the number of vehicles on each of the incoming and outgoing lanes at intersection $i$, the distance of the EMV to the intersection, the estimated time of arrival ($\mathsf{ETA}$), and which link the EMV will be routed to ($\mathsf{Next}$). Therefore, we have
\begin{displaymath}
s^t_i 
= 
\{x^t(l), x^t(m),  d^t_{\text{EMV}}[L_{ji}], \mathsf{ETA}^{t'}_i, \mathsf{Next}^{t'}_i \},
\end{displaymath}
where $L_{ji}$ denotes the links incoming to intersection $i$, and with a slight abuse of notation $l$ and $m$ denote the set of incoming and outgoing lanes, respectively. For primary pre-emption agents, one of the elements of $d^t_{\text{EMV}}$ represents the distance of the EMV to the intersection in the corresponding link and the rest of the elements are set to -1. For secondary pre-emption agents and normal agents, every element of $d^t_{\text{EMV}}$ is set to -1.
%
%
%
\\
\textbf{$\bullet$ Action:} We define the action of an agent as one of the 8 signal phases shown in Fig.~\ref{fig_phases}. This enables more flexible signal patterns as compared to the traditional cyclical patterns. Also, due to safety concerns, once a signal phase has been initiated, it remains unchanged for certain minimum amount of time.
\\
\textbf{$\bullet$ Reward:} PressLight has shown that minimizing the pressure can encourage efficient vehicle passage \cite{wei2019presslight}; here we adopt a similar approach for the normal agents. For secondary pre-emption agents, we additionally encourage less vehicle on the link where the EMV is about to enter in order to encourage efficient EMV passage. For primary pre-emption agents, we assign a unit penalty at each time step to encourage fast EMV passage. Thus, the local reward for agent $i$ at time $t$ is defined as
\begin{equation}
\label{eqn:reward}
r_{i}^{t} 
=
\left\{
\begin{array}{ll}
-P_{i}^{t} 
&\quad 
i \notin \{i_p, i_s\}\textrm{, i.e., Normal Agent},
\\
- \beta P_{i}^{t} - \frac{1-\beta}{|L_{i_pi}|}\sum\limits_{l\in L_{i_pi}} \frac{x(l)}{x_{max}(l)}  
&\quad 
i=i_s\textrm{, i.e., Secondary pre-emption agent,}
\\
-1 
&\quad 
i=i_p\textrm{, i.e., Primary pre-emption agent.}
\end{array}
\right.
\end{equation}

\subsection{Multi-agent Advantage Actor-critic}
To realize the decentalized RL framework, we adopt a multi-agent advantage actor-critic (MA2C) approach that is similar to \cite{chu2019multi}. In this formulation, the RL agent at intersection $i$ learns a policy $\pi_{\theta_i}$ (actor) and the corresponding value function $V_{\phi_i}$ (critic), where ${\theta_i}$ and ${\phi_i}$ are learnable neural network parameters. In our formulation, the local state includes dynamic routing information and the local reward encourages efficient passage of the EMV, whereas \cite{chu2019multi} focuses on a general traffic signal control problem with no EMV considered. Please see Section~S3 for additional details.
\section{Experiments and Results}
To evaluate performance of our proposed EMVLight framework, we carry out experiments using the \emph{SUMO} \cite{lopez2018microscopic} simulation tool that can simulate traffic dynamics in both microscopic and macroscopic settings. In our experiments, the decentralized RL agents collect observations from SUMO and the preferred signal phases are then fed back into the traffic simulator.

\subsection{Datasets and Maps Descriptions}
We carry out our  experiments using both synthetic and real-world maps.

\begin{figure}[b!]
\centering
\vspace{-2em}
\includegraphics[width=0.9\linewidth]{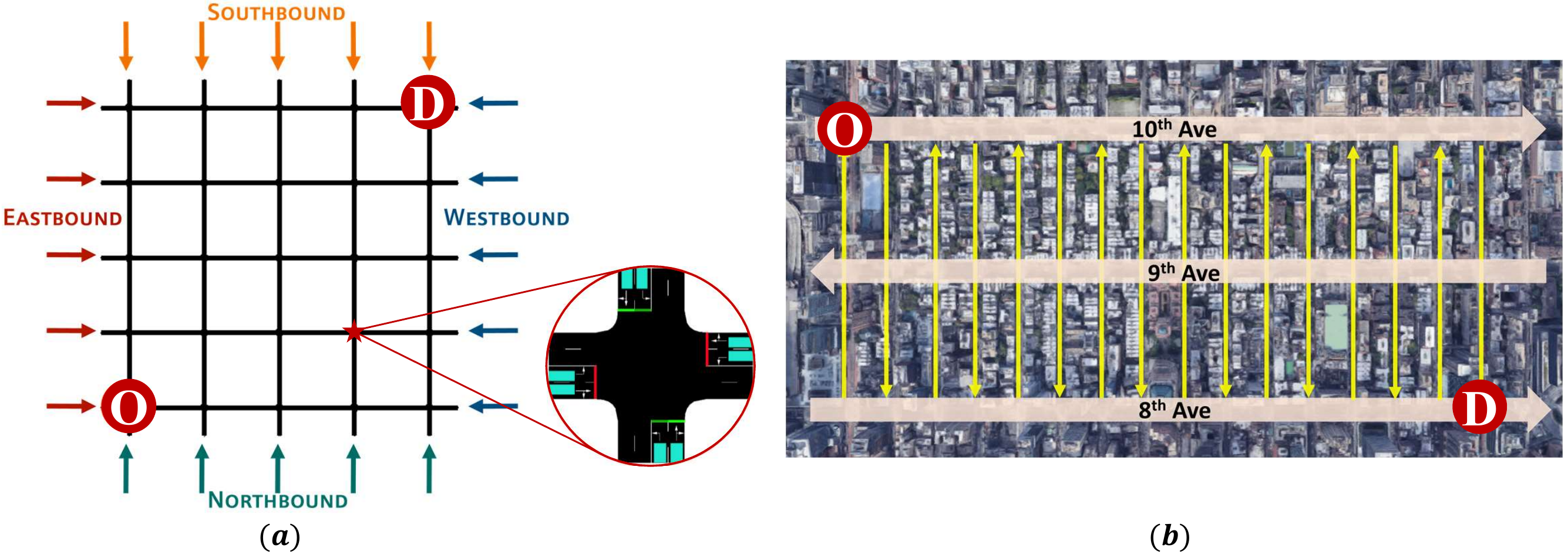}
\vspace{-0.75em}
\caption{\small{\emph{Left}: the synthetic $\text{grid}_{5\times 5}$. \emph{Right}: an intersection illustration in SUMO, the teal area are inductive loop detected area. Origin and destination for EMV are labeled. Manhattan map: a 16-by-3 traffic network in Hell's Kitchen area. Origin and destination for the EMV dispatching are labeled as ``\textbf{O}" and ``\textbf{D}", respectively.}}
\label{fig_synthetic_map}
\end{figure}

\textbf{Synthetic Grid:} We first consider a synthetic grid of size $5 \times 5$, where the intersections are connected with bi-directional links with each link containing two lanes (Fig.~\ref{fig_synthetic_map}a).
For the synthetic map, we consider two different traffic flow configurations with peak/non-peak flow values of 240/200 \emph{veh/lane/hr} and 320/120 \emph{veh/lane/hr}. In both configurations, vehicles enter the grid via north/south-bound links and leave the grid via east/west-bound links. The traffic for this map has a time span of 1200s and the EMV is dispatched at $t =600s$ to ensure roads are compacted before the EMV starts travelling; also, the peak flow occurs between 400s-800s.

\textbf{Manhattan Map:} This is a $16 \times 3$ traffic network extracted from Manhattan Hell's Kitchen area (Fig.~\ref{fig_synthetic_map}b). In this traffic network, intersections are connected by 16 one-way 2-lane streets and 3 one-way 4-lane avenues. The traffic flow for this map is generated from open-source NYC taxi data (\emph{https://traffic-signal-control.github.io/}).
%
%
%

\subsection{Baselines}
As RL-based approaches for efficient EMV passage do not yet exist, we select traditional methods and RL methods for individual subproblems and combine them to set up baselines. For traffic signal pre-emption, we use \emph{Walabi} \cite{bieker2019modelling}, a rule-based method that implements \emph{Green Wave} \cite{corman2009evaluation} for EMVs. We then integrate Walabi (\emph{W}) with combinations of \textit{routing} and \emph{traffic signal control} strategies. For routing, we consider 2 baselines -- a \emph{Static} routing based on A*; and a \emph{dynamic} routing by running A* every $50s$. For controlling the traffic signals, we consider the following 4 baseline approaches -- a cyclical \emph{fixed time (FT)} traffic phases with random offset that splits all phases with an predefined green ratio \cite{roess2004traffic}; a network-level signal control strategy based on \emph{max pressure (MP)} that smoothens congestion by aggressively selecting the phase with the maximum pressure \cite{varaiya2013max}; a Q-learning based \emph{coordinated learner (CL)} that directly learns joint local value functions for adjacent intersections \cite{van2016coordinated}; and finally \emph{PressLight (PL)} which uses RL to optimize the pressure at each intersection \cite{wei2019presslight}.

\subsection{Results}
We evaluate performance of models under two metrics: \emph{EMV travel time}, which reflects routing and pre-emption ability, and \emph{average travel time}, which indicates the ability of traffic signal control for efficient vehicle passage.  The performance of our EMVLight and the baselines in both the synthetic and the Manhattan map is shown in Table \ref{tab_performance_review}. The results of all methods are averaged over five independent runs and RL methods are tested with random seeds. We observe that EMVLight outperforms all baseline models under both metrics.
\begin{table*}[h!]
\vspace{-0.5em}
\centering
\fontsize{9.0pt}{10.0pt} \selectfont
\setlength\tabcolsep{3.4pt}
\begin{tabular}{ccccccc}
\hline
\multirow{2}{*}{Method} & \multicolumn{3}{c}{EMV Travel Time [s]}             & \multicolumn{3}{c}{Average Travel Time [s]}         \\ \cline{2-7} 
                        & Config 1        & Config 2        & Mahanttan       & Config 1        & Config 2        & Mahanttan       \\ \hline
No EMV + FT             & N/A             & N/A             & N/A             & 353.43          & 371.13          & 1649.64         \\ \hline
W + Static + FT         & 257.20          & 272.00          & 487.20          & 372.19          & 389.13          & 1711.03         \\
W + Static + MP         & 255.00          & 269.00          & 461.80          & 349.38          & 352.54          & 708.13          \\
W + Static + GRL        & 281.20          & 286.20          & 492.20          & 503.35          & 524.26          & 2013.54         \\
W + Static + PL         & 276.00          & 282.20          & 476.00          & 358.18          & 369.45          & 1410.76         \\ \hline
W + Dynamic + FT        & 229.60          & 231.20          & 442.60          & 370.09          & 293.40          & 1699.30         \\
W + Dynamic + MP        & 226.20          & 234.60          & 438.80          & 345.45          & 348.43          & 721.32          \\
W + Dynamic + GRL       & 273.40          & 269.60          & 450.20          & 514.29          & 536.78          & 1987.86         \\
W + Dynamic + PL        & 251.20          & 257.80          & 436.20          & 359.31          & 342.59          & 1412.12         \\ \hline
EMVLight                & \textbf{198.60} & \textbf{192.20} & \textbf{391.80} & \textbf{322.40} & \textbf{318.76} & \textbf{681.23} \\ \hline
\end{tabular}
\vspace{-0.25em}
\caption{\small{\emph{Performance comparison of different methods evaluated on four configurations in the synthetic traffic grid as well as Manhattan Map. For both metrics, the lower value indicates better performance. The lowest values are highlighted in bold. The average travel time of Manhattan map (1649.64) is retrieved from data.}}}
\vspace{-0.25em}
\label{tab_performance_review}
\end{table*}

In terms of EMV travel time $T_{\textrm{EMV}}$, the dynamic routing baseline performs better than static routing baselines. This is expected since dynamic routing considers the time-dependent nature of traffic conditions and update optimal route accordingly. EMVLight further reduces EMV travel time by 18\% in average as compared to dynamic routing baselines. This advantage in performance can be attributed to the design of secondary pre-emption agents. This type of agents learn to ``reserve a link" by choosing signal phases that help clear the vehicles in the link to encourage high speed EMV passage (Eqn.~\eqref{eqn:reward}).

As for average travel time $T_{\textrm{avg}}$, we first notice that the traditional pre-emption technique (W+Static+FT) indeed increases the average travel time by around 10\% as compared to a traditional Fix Time strategy without EMV (denoted as ``FT w/o EMV" in Table \ref{tab_performance_review}), thus decreasing the efficiency of vehicle passage. Different traffic signal control strategies have a direct impact on overall efficiency. Fixed Time is designed to handle steady traffic flow. Max Pressure, as a SOTA traditional method, outperforms Fix Time and, surprisingly, outperforms both RL baselines in terms of overall efficiency. This shows that pressure is an effective indicator for reducing congestion and this is why we incorporate pressure in our reward design. Coordinate Learner performs the worst probably because its reward is not based on pressure. PressLight doesn't beat Max Pressure because it has a reward design that focuses on smoothing vehicle densities along a major direction, e.g. an arterial. Grid networks with the presence of EMV make PressLight less effective. Our EMVLight improves its pressure-based reward design to encourage smoothing vehicle densities of all directions for each intersection. This enable us to achieve an advantage of 5\% over our best baselines (Max Pressure).
\section{Conclusion}
In this paper, we proposed a decentralized reinforcement learning framework, EMVLight, to facilitate the efficient passage of EMVs and reduce traffic congestion at the same time.
Leveraging the multi-agent A2C framework, agents incorporate dynamic routing and cooperatively control traffic signals to reduce EMV travel time and average travel time of non-EMVs. 
Evaluated on both synthetic and real-world map, EMVLight significantly outperforms the existing methods.
Future work will explore more realistic microscopic interaction between EMV and non-EMVs, efficient passage of multiple EMVs and closing the sim-to-real gap.

%
%
%
%
\newpage
{
\bibliography{main}

\begin{thebibliography}{10}
\providecommand{\url}[1]{#1}
\csname url@samestyle\endcsname
\providecommand{\newblock}{\relax}
\providecommand{\bibinfo}[2]{#2}
\providecommand{\BIBentrySTDinterwordspacing}{\spaceskip=0pt\relax}
\providecommand{\BIBentryALTinterwordstretchfactor}{4}
\providecommand{\BIBentryALTinterwordspacing}{\spaceskip=\fontdimen2\font plus
\BIBentryALTinterwordstretchfactor\fontdimen3\font minus
  \fontdimen4\font\relax}
\providecommand{\BIBforeignlanguage}[2]{{%
\expandafter\ifx\csname l@#1\endcsname\relax
\typeout{** WARNING: IEEEtran.bst: No hyphenation pattern has been}%
\typeout{** loaded for the language `#1'. Using the pattern for}%
\typeout{** the default language instead.}%
\else
\language=\csname l@#1\endcsname
\fi
#2}}
\providecommand{\BIBdecl}{\relax}
\BIBdecl

\bibitem{nichol2008regional}
G.~Nichol, E.~Thomas, C.~W. Callaway, J.~Hedges, J.~L. Powell, T.~P.
  Aufderheide, T.~Rea, R.~Lowe, T.~Brown, J.~Dreyer \emph{et~al.}, ``Regional
  variation in out-of-hospital cardiac arrest incidence and outcome,''
  \emph{Jama}, vol. 300, no.~12, pp. 1423--1431, 2008.

\bibitem{end-to-end-response-times}
N.~Analytics, ``End-to-end response times,''
  \url{https://www1.nyc.gov/site/fdny/about/resources/data-and-analytics/end-to-end-response-times.page},
  2021.

\bibitem{wei2019presslight}
H.~Wei, C.~Chen, G.~Zheng, K.~Wu, V.~Gayah, K.~Xu, and Z.~Li, ``Presslight:
  Learning max pressure control to coordinate traffic signals in arterial
  network,'' in \emph{Proceedings of the 25th ACM SIGKDD International
  Conference on Knowledge Discovery \& Data Mining}, 2019, pp. 1290--1298.

\bibitem{chu2019multi}
T.~Chu, J.~Wang, L.~Codec{\`a}, and Z.~Li, ``Multi-agent deep reinforcement
  learning for large-scale traffic signal control,'' \emph{IEEE Transactions on
  Intelligent Transportation Systems}, 2019.

\bibitem{lopez2018microscopic}
P.~A. Lopez, M.~Behrisch, L.~Bieker-Walz, J.~Erdmann, Y.-P. Fl{\"o}tter{\"o}d,
  R.~Hilbrich, L.~L{\"u}cken, J.~Rummel, P.~Wagner, and E.~Wie{\ss}ner,
  ``Microscopic traffic simulation using sumo,'' in \emph{2018 21st
  International Conference on Intelligent Transportation Systems (ITSC)}.\hskip
  1em plus 0.5em minus 0.4em\relax IEEE, 2018, pp. 2575--2582.

\bibitem{bieker2019modelling}
L.~Bieker-Walz and M.~Behrisch, ``Modelling green waves for emergency vehicles
  using connected traffic data,'' \emph{EPiC Series in Computing}, vol.~62, pp.
  1--11, 2019.

\bibitem{corman2009evaluation}
F.~Corman, A.~D’Ariano, D.~Pacciarelli, and M.~Pranzo, ``Evaluation of green
  wave policy in real-time railway traffic management,'' \emph{Transportation
  Research Part C: Emerging Technologies}, vol.~17, no.~6, pp. 607--616, 2009.

\bibitem{roess2004traffic}
R.~P. Roess, E.~S. Prassas, and W.~R. McShane, \emph{Traffic
  engineering}.\hskip 1em plus 0.5em minus 0.4em\relax Pearson/Prentice Hall,
  2004.

\bibitem{varaiya2013max}
P.~Varaiya, ``Max pressure control of a network of signalized intersections,''
  \emph{Transportation Research Part C: Emerging Technologies}, vol.~36, pp.
  177--195, 2013.

\bibitem{van2016coordinated}
E.~Van~der Pol and F.~A. Oliehoek, ``Coordinated deep reinforcement learners
  for traffic light control,'' \emph{Proceedings of Learning, Inference and
  Control of Multi-Agent Systems (at NIPS 2016)}, 2016.

\bibitem{wang2013development}
J.~Wang, W.~Ma, and X.~Yang, ``Development of degree-of-priority based control
  strategy for emergency vehicle preemption operation,'' \emph{Discrete
  dynamics in nature and society}, vol. 2013, 2013.

\bibitem{Mu2018Route}
H.~Mu, Y.~Song, and L.~Liu, ``Route-based signal preemption control of
  emergency vehicle,'' \emph{Journal of Control Science and Engineering}, vol.
  2018, pp. 1--11, 04 2018.

\bibitem{kwon2003route}
E.~Kwon, S.~Kim, and R.~Betts, ``Route-based dynamic preemption of traffic
  signals for emergency vehicle operations,'' in \emph{Transportation Research
  Board 82nd Annual MeetingTransportation Research Board}, 2003.

\bibitem{JOTSHI20091}
A.~Jotshi, Q.~Gong, and R.~Batta, ``Dispatching and routing of emergency
  vehicles in disaster mitigation using data fusion,'' \emph{Socio-Economic
  Planning Sciences}, vol.~43, no.~1, pp. 1 -- 24, 2009.

\bibitem{nordin2012finding}
N.~A.~M. Nordin, Z.~A. Zaharudin, M.~A. Maasar, and N.~A. Nordin, ``Finding
  shortest path of the ambulance routing: Interface of a-star algorithm using c
  programming,'' in \emph{2012 IEEE Symposium on Humanities, Science and
  Engineering Research}.\hskip 1em plus 0.5em minus 0.4em\relax IEEE, 2012, pp.
  1569--1573.

\bibitem{ziliaskopoulos1993time}
A.~K. Ziliaskopoulos and H.~S. Mahmassani, ``Time-dependent, shortest-path
  algorithm for real-time intelligent vehicle highway system applications,'' in
  \emph{Transportation Research Record}, 1993, pp. 94--100.

\bibitem{musolino2013travel}
G.~Musolino, A.~Polimeni, C.~Rindone, and A.~Vitetta, ``Travel time forecasting
  and dynamic routes design for emergency vehicles,'' \emph{Procedia-Social and
  Behavioral Sciences}, vol.~87, pp. 193--202, 2013.

\bibitem{haghani2003optimization}
A.~Haghani, H.~Hu, and Q.~Tian, ``An optimization model for real-time emergency
  vehicle dispatching and routing,'' in \emph{82nd annual meeting of the
  Transportation Research Board, Washington, DC}.\hskip 1em plus 0.5em minus
  0.4em\relax Citeseer, 2003.

\bibitem{koh2020real}
S.~Koh, B.~Zhou, H.~Fang, P.~Yang, Z.~Yang, Q.~Yang, L.~Guan, and Z.~Ji,
  ``Real-time deep reinforcement learning based vehicle navigation,''
  \emph{Applied Soft Computing}, vol.~96, p. 106694, 2020.

\bibitem{Asaduzzaman2017APriority}
M.~{Asaduzzaman} and K.~{Vidyasankar}, ``A priority algorithm to control the
  traffic signal for emergency vehicles,'' in \emph{2017 IEEE 86th Vehicular
  Technology Conference (VTC-Fall)}, 2017, pp. 1--7.

\bibitem{Lu2019Literature}
L.~{Lu} and S.~{Wang}, ``Literature review of analytical models on emergency
  vehicle service: Location, dispatching, routing and preemption control,'' in
  \emph{2019 IEEE Intelligent Transportation Systems Conference (ITSC)}, 2019,
  pp. 3031--3036.

\bibitem{humagain2020systematic}
S.~Humagain, R.~Sinha, E.~Lai, and P.~Ranjitkar, ``A systematic review of route
  optimisation and pre-emption methods for emergency vehicles,''
  \emph{Transport reviews}, vol.~40, no.~1, pp. 35--53, 2020.

\bibitem{you2018structurally}
C.~You, Q.~Yang, L.~Gjesteby, G.~Li, S.~Ju, Z.~Zhang, Z.~Zhao, Y.~Zhang,
  W.~Cong, G.~Wang \emph{et~al.}, ``Structurally-sensitive multi-scale deep
  neural network for low-dose {CT} denoising,'' \emph{IEEE Access}, 2018.

\bibitem{cheng2016hybrid}
L.~Cheng and C.~You, ``Hybrid non-linear dimensionality reduction method
  framework based on random projections,'' in \emph{2016 IEEE International
  Conference on Cloud Computing and Big Data Analysis (ICCCBDA)}.\hskip 1em
  plus 0.5em minus 0.4em\relax IEEE, 2016, pp. 43--48.

\bibitem{li2019novel}
G.~Li, S.~Luo, C.~You, M.~Getzin, L.~Zheng, G.~Wang, and N.~Gu, ``A novel
  calibration method incorporating nonlinear optimization and ball-bearing
  markers for cone-beam ct with a parameterized trajectory,'' \emph{Medical
  physics}, 2019.

\bibitem{cheng2016random}
L.~Cheng, C.~You, and Y.~Guan, ``Random projections for non-linear
  dimensionality reduction,'' \emph{International Journal of Machine Learning
  and Computing}, vol.~6, no.~4, pp. 220--225, 2016.

\bibitem{cheng2017body}
L.~Cheng, C.~You, Y.~Guan, and Y.~Yu, ``Body activity recognition using
  wearable sensors,'' in \emph{2017 Computing Conference}.\hskip 1em plus 0.5em
  minus 0.4em\relax IEEE, 2017, pp. 756--765.

\bibitem{you2019low}
C.~You, L.~Yang, Y.~Zhang, and G.~Wang, ``Low-{D}ose {CT} via {D}eep {CNN} with
  {S}kip {C}onnection and {N}etwork in {N}etwork,'' in \emph{Developments in
  X-Ray Tomography XII}.\hskip 1em plus 0.5em minus 0.4em\relax International
  Society for Optics and Photonics, 2019.

\bibitem{you2019ct}
C.~You, G.~Li, Y.~Zhang, X.~Zhang, H.~Shan, M.~Li, S.~Ju, Z.~Zhao, Z.~Zhang,
  W.~Cong \emph{et~al.}, ``{CT} super-resolution {GAN} constrained by the
  identical, residual, and cycle learning ensemble (gan-circle),'' \emph{IEEE
  Transactions on Medical Imaging}, 2019.

\bibitem{lyu2019super}
Q.~Lyu, C.~You, H.~Shan, Y.~Zhang, and G.~Wang, ``Super-resolution mri and ct
  through gan-circle,'' in \emph{Developments in X-ray tomography XII}.\hskip
  1em plus 0.5em minus 0.4em\relax International Society for Optics and
  Photonics, 2019.

\bibitem{you2020unsupervised}
C.~You, J.~Yang, J.~Chapiro, and J.~S. Duncan, ``Unsupervised wasserstein
  distance guided domain adaptation for 3d multi-domain liver segmentation,''
  in \emph{Interpretable and Annotation-Efficient Learning for Medical Image
  Computing}, 2020.

\bibitem{you2020contextualized}
C.~You, N.~Chen, and Y.~Zou, ``Contextualized attention-based knowledge
  transfer for spoken conversational question answering,'' in
  \emph{INTERSPEECH}, 2021.

\bibitem{you2020data}
C.~You, N.~Chen, F.~Liu, D.~Yang, and Y.~Zou, ``Towards data distillation for
  end-to-end spoken conversational question answering,'' \emph{arXiv preprint
  arXiv:2010.08923}, 2020.

\bibitem{yang2020nuset}
L.~Yang, R.~P. Ghosh, J.~M. Franklin, S.~Chen, C.~You, R.~R. Narayan, M.~L.
  Melcher, and J.~T. Liphardt, ``Nuset: A deep learning tool for reliably
  separating and analyzing crowded cells,'' \emph{PLoS computational biology},
  2020.

\bibitem{chen2020adaptive}
N.~Chen, F.~Liu, C.~You, P.~Zhou, and Y.~Zou, ``Adaptive bi-directional
  attention: Exploring multi-granularity representations for machine reading
  comprehension,'' in \emph{ICASSP}, 2020.

\bibitem{chen2021self}
N.~Chen, C.~You, and Y.~Zou, ``Self-supervised dialogue learning for spoken
  conversational question answering,'' in \emph{INTERSPEECH}, 2021.

\bibitem{you2021knowledge}
C.~You, N.~Chen, and Y.~Zou, ``Knowledge distillation for improved accuracy in
  spoken question answering,'' in \emph{ICASSP}, 2021.

\bibitem{you2021simcvd}
C.~You, Y.~Zhou, R.~Zhao, L.~Staib, and J.~S. Duncan, ``Simcvd: Simple
  contrastive voxel-wise representation distillation for semi-supervised
  medical image segmentation,'' \emph{arXiv preprint arXiv:2108.06227}, 2021.

\bibitem{you2021mrd}
C.~You, N.~Chen, and Y.~Zou, ``{MRD-N}et: {M}ulti-{M}odal {R}esidual
  {K}nowledge {D}istillation for {S}poken {Q}uestion {A}nswering,'' in
  \emph{{IJCAI}}, 2021.

\bibitem{you2021self}
------, ``Self-supervised contrastive cross-modality representation learning
  for spoken question answering,'' in \emph{Findings of the Association for
  Computational Linguistics: EMNLP}, 2021.

\bibitem{you2022megan}
C.~You, L.~Han, A.~Feng, R.~Zhao, H.~Tang, and W.~Fan, ``Megan: Memory enhanced
  graph attention network for space-time video super-resolution,'' in
  \emph{WACV}, 2022.

\bibitem{xu2021semantic}
W.~Xu, P.~Zhou, C.~You, and Y.~Zou, ``Semantic transportation prototypical
  network for few-shot intent detection,'' in \emph{INTERSPEECH}, 2021.

\bibitem{liu2021aligning}
F.~Liu, X.~Wu, C.~You, S.~Ge, Y.~Zou, and X.~Sun, ``Aligning source visual and
  target language domains for unpaired video captioning,'' \emph{IEEE
  Transactions on Pattern Analysis and Machine Intelligence}, 2021.

\bibitem{liu2021auto}
F.~Liu, C.~You, X.~Wu, S.~Ge, X.~Sun \emph{et~al.}, ``Auto-encoding knowledge
  graph for unsupervised medical report generation,'' in \emph{NeurIPS}, 2021.

\bibitem{you2022class}
C.~You, R.~Zhao, F.~Liu, S.~Chinchali, U.~Topcu, L.~Staib, and J.~S. Duncan,
  ``Class-aware generative adversarial transformers for medical image
  segmentation,'' \emph{arXiv preprint arXiv:2201.10737}, 2022.

\bibitem{abdulhai2003reinforcement}
B.~Abdulhai, R.~Pringle, and G.~J. Karakoulas, ``Reinforcement learning for
  true adaptive traffic signal control,'' \emph{Journal of Transportation
  Engineering}, vol. 129, no.~3, pp. 278--285, 2003.

\bibitem{prashanth2010reinforcement}
L.~Prashanth and S.~Bhatnagar, ``Reinforcement learning with function
  approximation for traffic signal control,'' \emph{IEEE Transactions on
  Intelligent Transportation Systems}, vol.~12, no.~2, pp. 412--421, 2010.

\bibitem{wei2019colight}
H.~Wei, N.~Xu, H.~Zhang, G.~Zheng, X.~Zang, C.~Chen, W.~Zhang, Y.~Zhu, K.~Xu,
  and Z.~Li, ``Colight: Learning network-level cooperation for traffic signal
  control,'' in \emph{Proceedings of the 28th ACM International Conference on
  Information and Knowledge Management}, 2019, pp. 1913--1922.

\bibitem{zheng2019frap}
G.~Zheng, Y.~Xiong, X.~Zang, J.~Feng, H.~Wei, H.~Zhang, Y.~Li, K.~Xu, and
  Z.~Li, ``Learning phase competition for traffic signal control,'' in
  \emph{Proceedings of the 28th ACM International Conference on Information and
  Knowledge Management}, 2019, pp. 1963--1972.

\bibitem{ThousandLights}
\BIBentryALTinterwordspacing
C.~Chen, H.~Wei, N.~Xu, G.~Zheng, M.~Yang, Y.~Xiong, K.~Xu, and Z.~Li, ``Toward
  a thousand lights: Decentralized deep reinforcement learning for large-scale
  traffic signal control,'' \emph{Proceedings of the AAAI Conference on
  Artificial Intelligence}, vol.~34, no.~04, pp. 3414--3421, Apr. 2020.
  [Online]. Available:
  \url{https://ojs.aaai.org/index.php/AAAI/article/view/5744}
\BIBentrySTDinterwordspacing

\bibitem{Zang_Yao_Zheng_Xu_Xu_Li_2020}
X.~Zang, H.~Yao, G.~Zheng, N.~Xu, K.~Xu, and Z.~Li, ``Metalight: Value-based
  meta-reinforcement learning for traffic signal control,'' \emph{Proceedings
  of the AAAI Conference on Artificial Intelligence}, vol.~34, no.~01, pp.
  1153--1160, Apr. 2020.

\bibitem{el2013multiagent}
S.~El-Tantawy, B.~Abdulhai, and H.~Abdelgawad, ``Multiagent reinforcement
  learning for integrated network of adaptive traffic signal controllers
  (marlin-atsc): methodology and large-scale application on downtown toronto,''
  \emph{IEEE Transactions on Intelligent Transportation Systems}, vol.~14,
  no.~3, pp. 1140--1150, 2013.

\bibitem{aslani2017adaptive}
M.~Aslani, M.~S. Mesgari, and M.~Wiering, ``Adaptive traffic signal control
  with actor-critic methods in a real-world traffic network with different
  traffic disruption events,'' \emph{Transportation Research Part C: Emerging
  Technologies}, vol.~85, pp. 732--752, 2017.

\bibitem{xu2021hierarchically}
B.~Xu, Y.~Wang, Z.~Wang, H.~Jia, and Z.~Lu, ``Hierarchically and cooperatively
  learning traffic signal control,'' in \emph{Proceedings of the AAAI
  Conference on Artificial Intelligence}, vol.~35, 2021, pp. 669--677.

\bibitem{wei2019survey}
H.~Wei, G.~Zheng, V.~Gayah, and Z.~Li, ``A survey on traffic signal control
  methods,'' \emph{arXiv preprint arXiv:1904.08117}, 2019.

\bibitem{gajda2001vehicle}
J.~Gajda, R.~Sroka, M.~Stencel, A.~Wajda, and T.~Zeglen, ``A vehicle
  classification based on inductive loop detectors,'' in \emph{IMTC 2001.
  Proceedings of the 18th IEEE Instrumentation and Measurement Technology
  Conference. Rediscovering Measurement in the Age of Informatics},
  vol.~1.\hskip 1em plus 0.5em minus 0.4em\relax IEEE, 2001, pp. 460--464.

\bibitem{buchenscheit2009vanet}
A.~Buchenscheit, F.~Schaub, F.~Kargl, and M.~Weber, ``A vanet-based emergency
  vehicle warning system,'' in \emph{2009 IEEE Vehicular Networking Conference
  (VNC)}.\hskip 1em plus 0.5em minus 0.4em\relax IEEE, 2009, pp. 1--8.

\bibitem{wang2013design}
Y.~Wang, Z.~Wu, X.~Yang, and L.~Huang, ``Design and implementation of an
  emergency vehicle signal preemption system based on cooperative
  vehicle-infrastructure technology,'' \emph{Advances in Mechanical
  Engineering}, vol.~5, p. 834976, 2013.

\bibitem{noori2016connected}
H.~Noori, L.~Fu, and S.~Shiravi, ``A connected vehicle based traffic signal
  control strategy for emergency vehicle preemption,'' in \emph{Transportation
  Research Board 95th Annual Meeting, Report Np. 16-6763}, 2016.

\bibitem{su2021hybrid}
H.~Su, Z.~Ji, K.~Johansson, L.~Jin \emph{et~al.}, ``A hybrid queuing model for
  coordinated vehicle platooning on mixed-autonomy highways: Training and
  validation,'' \emph{arXiv preprint arXiv:2103.14202}, 2021.

\bibitem{foerster2017stabilising}
J.~Foerster, N.~Nardelli, G.~Farquhar, T.~Afouras, P.~H. Torr, P.~Kohli, and
  S.~Whiteson, ``Stabilising experience replay for deep multi-agent
  reinforcement learning,'' in \emph{International conference on machine
  learning}.\hskip 1em plus 0.5em minus 0.4em\relax PMLR, 2017, pp. 1146--1155.

\bibitem{su2021dynamic}
H.~Su, K.~Shi, J.~Y.~J. Chow, and L.~Jin, ``Dynamic queue-jump lane for
  emergency vehicles under partially connected settings: A multi-agent deep
  reinforcement learning approach,'' 2021.

\end{thebibliography}
\bibliographystyle{IEEEtran}
}

\newpage
\appendix
\pagebreak
\renewcommand\appendixpagename{Supplementary Material}
\appendix
\appendixpage
\renewcommand{\thesection}{S\arabic{section}}

\section{Related Work:}
\textbf{Conventional routing optimization and traffic signal pre-emption for EMVs.}
%
%
Although, in reality, routing and pre-emption are coupled, the existing methods usually solve them separately. Many of the existing approaches leverage Dijkstra's shortest path algorithm to get the optimal route \cite{wang2013development, Mu2018Route, kwon2003route, JOTSHI20091}. An A* algorithm for ambulance routing has been proposed by \cite{nordin2012finding}. However, as this line of work assumes that the routes and traffic conditions are fixed and static, they fail to address the dynamic nature of real-world traffic flows. Another line of work has considered the change of traffic flows over time. \cite{ziliaskopoulos1993time} has proposed a shortest-path algorithm for time-dependent traffic networks, but the travel time associated with each edge at each time step is assumed to be known in prior. \cite{musolino2013travel} proposes different routing strategies for different times in a day (e.g., peak/non-peak hours) based on traffic history data at those times. However, in the problem of our consideration, routing and pre-emption strategies can significantly affect the travel time associated with each edge during the EMV passage, and the existing methods cannot deal with this kind of real-time changes. \cite{haghani2003optimization} formulated the dynamic shortest path problem as a mixed-integer programming problem. \cite{koh2020real} has used RL for real-time vehicle navigation and routing. However, both of these studies have tackled a general routing problem, and signal pre-emption and its influence on traffic have not been modeled. Once an optimal route for the EMV has been determined, traffic signal pre-emption is deployed. A common pre-emption strategy is to extend the green phases of green lights to let the EMV pass each intersection along a fixed optimal route \cite{wang2013development, bieker2019modelling}. \cite{Asaduzzaman2017APriority} has proposed pre-emption strategies for multiple EMV requests. Please refer to \cite{Lu2019Literature} and \cite{humagain2020systematic} for a thorough survey of conventional routing optimization and traffic signal pre-emption methods. We would also like to point out that the conventional methods prioritize EMV passage and have significant disturbances on the traffic flow which increases the average non-EMV travel time.
%
%
%
%

\textbf{RL-based traffic signal control.} 
Traffic signal pre-emption only alters the traffic phases at the intersections where an EMV travels through. However, to reduce congestion, traffic phases at nearby intersections also need to be changed cooperatively. Recently, deep neural network has the promise to revolutionize the real world applications, especially in the context of healthcare and humanitarian assistance \cite{you2018structurally,cheng2016hybrid,li2019novel,cheng2016random,cheng2017body,you2019low,you2019ct,lyu2019super,you2020unsupervised,you2020contextualized,you2020data,yang2020nuset,chen2020adaptive,chen2021self,you2021knowledge,you2021simcvd,you2021mrd,you2021self,you2022megan,xu2021semantic,liu2021aligning,liu2021auto,you2022class}. The coordination of traffic signals to mitigate traffic congestion is referred to as traffic signal control which has been addressed by leveraging deep RL in a growing body of work. Many of the existing approaches use Q-learning \cite{abdulhai2003reinforcement, prashanth2010reinforcement, wei2019presslight, wei2019colight, zheng2019frap, ThousandLights}. \cite{Zang_Yao_Zheng_Xu_Xu_Li_2020} leverages meta-learning algorithms to speed up Q-learning for traffic signal control. Another line of work has used actor-critic algorithms for traffic signal control \cite{el2013multiagent, aslani2017adaptive, chu2019multi}. \cite{xu2021hierarchically} proposes a hierarchical actor-critic method to encourage cooperation between intersections. Please refer to \cite{wei2019survey} for a review on traffic signal control methods. However, these RL-based traffic control methods focus on reducing the congestion in the traffic network and are not designed for EMV pre-emption. In contrast, our RL framework is built upon state-of-the-art ideas such as max pressure and is designed to reduce both EMV travel time and overall congestion. 

\section{Justification of the Agent Design}
The quantities in local agent state can be obtained at each intersection using various technologies. Numbers of vehicles on each lane $(x^t(l), x^t(m))$ can be obtained by vehicle detection technologies, such as inductive loop \cite{gajda2001vehicle} based on the hardware installed underground. The distance of the EMV to the intersection $d^t_{EMV}[L_{ji}]$ can be obtained by \emph{vehicle-to-infrastructure} technologies such as VANET \cite{buchenscheit2009vanet}, which broadcasts the real-time position of a vehicle to an intersection. Prior work \cite{wang2013design, noori2016connected, su2021hybrid} has explored these technologies for traffic signal pre-emption. 

The dynamic routing algorithm (Algorithm~1) can provide $(\mathsf{ETA}, \mathsf{Next})$ for each agent at every time step. However, due to the stochastic nature of traffic flows, updating the route too frequently might confuse the EMV driver, since the driver might be instructed a new route, say, every 5 seconds. 
There are many ways to ensure reasonable frequency. One option is to inform the driver only once while the EMV is travels in a single link. We implement it by updating the state of an RL agent $(\mathsf{ETA}^{t'}_i, \mathsf{Next}^{t'}_i)$ at the time step when the EMV travels through half of a link. For example, if the EMV travels through a link to agent $i$ from time step 11 to 20 in constant speed, then dynamic routing information in $s_i^{16}$ to $s_i^{20}$ are the same, which is $(\mathsf{ETA}_i^{15}, \mathsf{Next}_i^{15})$, i.e., $t'=15$.

As for the reward design, one might wonder how an agent can know its type. As we assume an agent can observe the state of its neighbors, agent type can be inferred from the observation. This will become clearer below.
\section{Multi-agent Actor Critic Method}\label{sec_multi_agent}
\textbf{Local Observation.} In an ideal setting, agents can observe the states of every other agent and leverage this global information to make a decision. However, this is not practical in our problem due to communication latency and will cause scalability issues. We assume agents can observe its own state and the states of its neighbors, i.e., $s^t_{\mathcal{V}_i} = \{s^t_j|j\in \mathcal{V}_i\}$. The agents feed this observation to its policy network $\pi_{\theta_i}$ and value network $V_{\phi_i}$.


\textbf{Fingerprint.} In multi-agent training, each agent treats other agents as part of the environment, but the policy of other agents are changing over time. \cite{foerster2017stabilising} introduces \emph{fingerprints} to inform agents about the changing policies of neighboring agents in multi-agent Q-learning. \cite{chu2019multi} brings fingerprints into MA2C. Here we use the probability simplex of neighboring policies $\pi^{t-1}_{\mathcal{N}_i} = \{\pi^{t-1}_j|j\in \mathcal{N}_i\}$ as fingerprints, and include it into the input of policy network and value network. Thus, our policy network can be written as $\pi_{\theta_i}(a_i^t|s^t_{\mathcal{V}_i}, \pi^{t-1}_{\mathcal{N}_i})$ and value network as $V_{\phi_i}(\Tilde{s}^t_{\mathcal{V}_i}, \pi^{t-1}_{\mathcal{N}_i})$, where $\Tilde{s}^t_{\mathcal{V}_i}$ is the local observation with spatial discount factor, which is introduced below. 

\textbf{Spatial Discount Factor and Adjusted Reward.} MA2C agents cooperatively optimize a global cumulative reward. We assume the global reward is decomposable as $r_t = \sum_{i\in \mathcal{V}} r^t_i$, where $r^t_i$ is defined in Eqn.~1. Instead of optimizing the same global reward for every agent, \cite{chu2019multi} proposes a spatial discount factor $\alpha$ to let each agent pay less attention to rewards of agents far away. The adjusted reward for agent $i$ is 
\begin{equation}
    \Tilde{r}_i^t = \sum_{d=0}^{D_i}\Big( \sum_{j\in\mathcal{V}|d(i, j)=d} (\alpha)^d r^t_j\Big),
\end{equation}
where $D_i$ is the maximum distance of agents in the graph from agent $i$. When $\alpha > 0$, the adjusted reward include global information, it seems this is in contradiction to the local communication assumption. However, since reward is only used for offline training, global reward information is allowed. Once trained, the RL agents can control traffic signal without relying on global information. 

\textbf{Temporal Discount Factor and Return.} 
The local return $\Tilde{R}^t_i$ is defined as the cumulative adjusted reward $\Tilde{R}^t_i := \sum_{\tau=t}^T \gamma^{\tau-t} \Tilde{r}^\tau_i$, where $\gamma$ is the temporal discount factor and $T$ is the length of an episode. we can estimate the local return using value function,
\begin{equation}
    \Tilde{R}^t_i = \Tilde{r}^t_i + \gamma V_{\phi_i^-}(\Tilde{s}^{t+1}_{\mathcal{V}_i}, \pi^{t}_{\mathcal{N}_i}|\pi_{\theta_{-i}^-}),
\end{equation}
where $\phi_i^-$ means parameters $\phi_i$ are frozen and $\theta_{-i}^-$ means the parameters of policy networks of all other agents are frozen. 

\textbf{Network architecture and training.} As traffic flow data are spatial temporal, we leverage a long-short term memory (LSTM) layer along with fully connected (FC) layers for policy network (actor) and value network (critic).  Our multi-agent actor-critic training pipeline is similar to that in \cite{chu2019multi}. We provide neural architecture details, policy loss expression,  value loss expression as well as a training pseudocode.

\section{Training Details}

\subsection{Value loss function}
With a batch of data $B = \{(s_i^t, \pi_i^t, a_i^t, s_i^{t+1}, r_i^t)_{i\in \mathcal{V}}^{t\in \mathcal{T}}\}$, each agent's value network is trained by minimizing the difference between bootstrapped estimated value and neural network approximated value
\begin{equation}
    \label{eqn:L_v}
    \mathcal{L}_v(\phi_i) = \frac{1}{2|B|} \sum_{B}\Big( \Tilde{R}^t_i - V_{\phi_i}(\Tilde{s}^t_{\mathcal{V}_i}, \pi^{t-1}_{\mathcal{N}_i}) \Big)^2.
\end{equation}

\subsection{Policy loss function}
Each agent's policy network is trained by minimizing its policy loss
\begin{align}
    \label{eqn:L_p}
    \mathcal{L}_p(\theta_i) = -& \frac{1}{|B|}\sum_{B} \bigg(\log \pi_{\theta_i}(a_i^t|s^t_{\mathcal{V}_i}, \pi^{t-1}_{\mathcal{N}_i}) \Tilde{A}^t_i \\
    &- \lambda \sum_{a_i \in \mathcal{A}_i} \pi_{\theta_i} \log \pi_{\theta_i} (a_i | s^t_{\mathcal{V}_i}, \pi^{t-1}_{\mathcal{N}_i}) \bigg),
\end{align}
where $\Tilde{A}^t_i = \Tilde{R}^t_i - V_{\phi_i^-}(\Tilde{s}^t_{\mathcal{V}_i}, \pi^{t-1}_{\mathcal{N}_i})$ is the estimated advantage which measures how much better the action $a^t_i$ is as compared to the average performance of the policy $\pi_{\theta_i}$ in the state $s_i^t$. The second term is a regularization term that encourage initial exploration, where $\mathcal{A}_i$ is the action set of agent $i$. For an intersection as shown in Fig. 1, $\mathcal{A}_i$ contains 8 traffic signal phases.

\subsection{Training algorithm}
Algorithm \ref{alg:training} shows the multi-agent A2C training process. %
\setcounter{algocf}{0}
\renewcommand{\thealgocf}{S\arabic{algocf}}
\begin{algorithm}[h]
    \caption{Multi-agent A2C Training}
    \label{alg:training}
    \SetEndCharOfAlgoLine{}
    \SetKwInOut{Input}{Input}
    \SetKwInOut{Output}{Output}
    \SetKwData{ETA}{ETA}
    \SetKwData{Next}{Next}
    \SetKwFor{ParrallelForEach}{foreach}{do (in parallel)}{endfor}
    \Input{\\\hspace{-3.7em}
        \begin{tabular}[t]{l @{\hspace{3.3em}} l}
        $T$ & maximum time step of an episode \\
        $N_{\mathrm{bs}}$ & batch size \\
        $\eta_\theta$  & learning rate for policy networks \\
        $\eta_\phi$    & learning rate for value networks \\
        $\alpha$       & spatial discount factor \\
        $\gamma$       & (temporal) discount factor\\ $\lambda$      & regularizer coefficient
        \end{tabular}
    }
    \Output{\\\hspace{-3.7em}
        \begin{tabular}[t]{l @{\hspace{1.4em}} l}
        $\{\phi_i\}_{i\in\mathcal{V}}$ & learned parameters in value networks \\
        $\{\theta_i\}_{i\in\mathcal{V}}$ & learned parameters in policy networks \\
        \end{tabular}
    }
    \textbf{initialize} $\{\phi_i\}_{i\in\mathcal{V}}$, $\{\theta_i\}_{i\in\mathcal{V}}$, $k \gets 0$, $B \gets \varnothing$;
    \textbf{initialize} SUMO, $t \gets 0$, \textbf{get} $\{s^0_i\}_{i\in\mathcal{V}}$\;
    \Repeat{Convergence}{
        \tcc{generate trajectories}
        \ParrallelForEach{$i \in \mathcal{V}$}{
            \textbf{sample} $a^t_i$ from $\pi^t_i$\;
            \textbf{receive} $\Tilde{r}^t_i$ and $s^{t+1}_i$\;
        }
        $B \gets B \cup \{(s_i^t, \pi_i^t, a_i^t, s_i^{t+1}, r_i^t)_{i\in \mathcal{V}}\}$\;
        $t \gets t+1$, $k \gets k+1$\;
        \If{$t == T$}{
            \textbf{initialize} SUMO, $t \gets 0$, \textbf{get} $\{s^0_i\}_{i\in\mathcal{V}}$\;
        }
        \tcc{update actors and critics}
        \If{$k == N_{\mathrm{bs}}$}{
            \ParrallelForEach{$i \in \mathcal{V}$}{
                \textbf{calculate} $\Tilde{r}^t_i$ (Eqn. (4)), $\Tilde{R}^t_i$ (Eqn. (5))\;
                $\phi_i \gets \phi_i - \eta_\phi \nabla \mathcal{L}_v(\phi_i)$\;
                $\theta_i \gets \theta_i - \eta_\theta \nabla \mathcal{L}_p(\theta_i)$\;
            }
            $k \gets 0, B \gets \varnothing$\;
        }
    }
    
\end{algorithm}
\section{Intralink EMV travel time} \label{sec_intralink}

The intra-link traffic pattern with the presence of an EMV on duty is complicated and is under-explored in the current literature. For simplicity, here we demonstrate a simple intra-link traffic model for a link with 2 lanes. The model can be easily extended for multiple lanes.

In a two-lane link, the EMV takes a lane and the non-EMVs on the other lane usually slows down or entirely stop. Some non-EMVs ahead of the EMV find pull-over spots in the other lane and park there. Those that cannot find a parking spot continue to drive in front of the EMV, potentially blocking the EMV passage \cite{su2021dynamic}. In this study, we propose a meso-scopic model to estimate the intra-link travel time of an EMV.
\begin{figure}[h]
    \centering
    \includegraphics[width=0.75\linewidth]{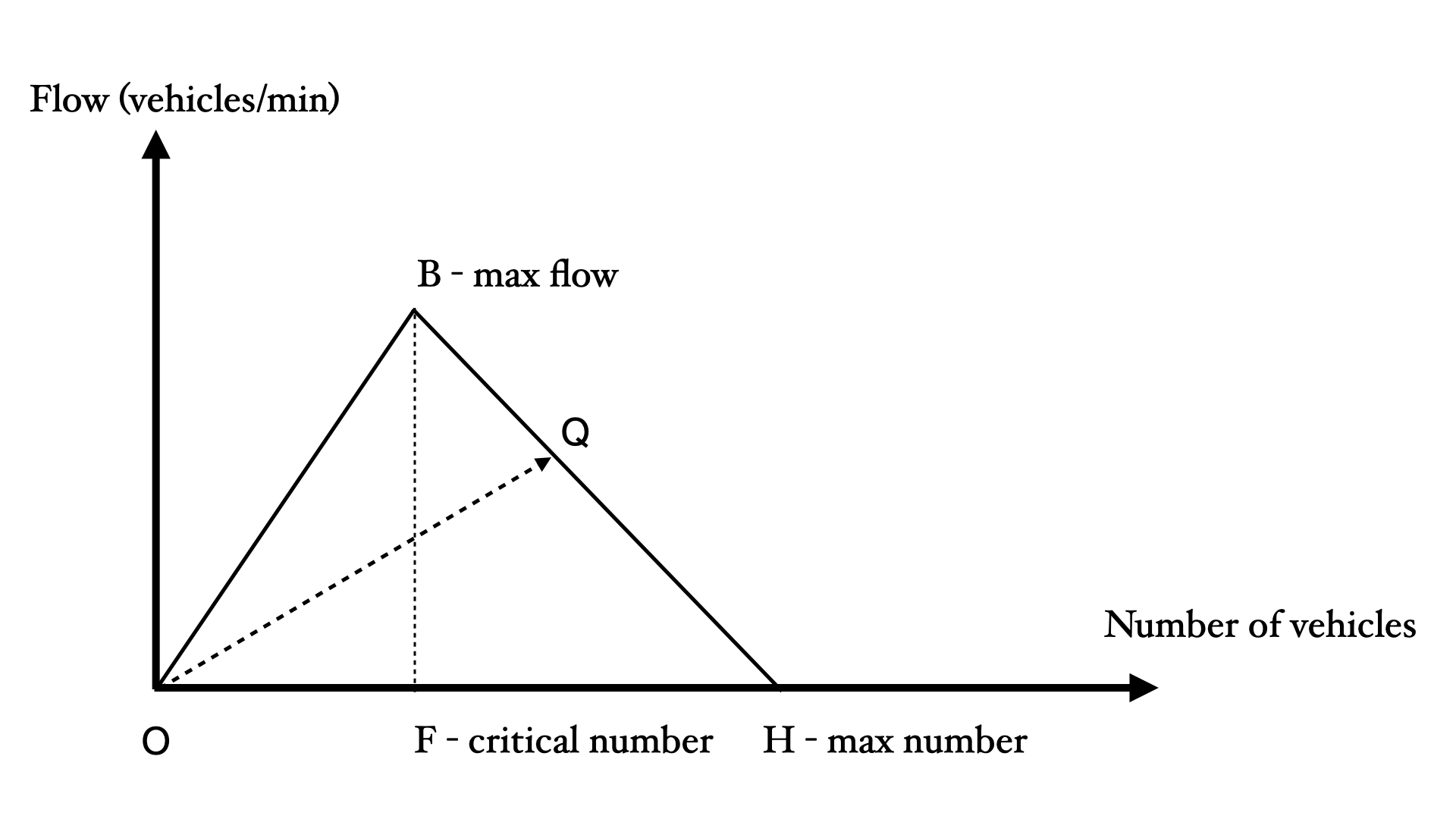}
  \caption{Normal traffic state.}
  \label{fig_two_lanes}
\end{figure}
Normally, the traffic flow of a link is modeled by a fundamental diagram, see Fig \ref{fig_two_lanes}. This diagram depicts a simplified relationship between the flow rate, i.e. number of vehicles passing within the unit amount of the time, and number of vehicles on a link. The max number of vehicles $H$ indicates the capacity of this link. The critical number of vehicles $F$ indicates the boundary differentiates the non-congested state and congested state. When the number of vehicles is smaller than $F$, all vehicles are traveling at the free flow speed, which is represented by the slope of $OB$. When number of vehicles is larger than $F$, vehicles are slowing down and traffic flows declines since the link is now congested. The max flow is attained when the number of vehicles is at $F$. The travel speed of vehicles in a congested state $Q$ is obtained by the slope of $OQ$.
\begin{figure}[h]
    \centering
    \includegraphics[width=0.75\linewidth]{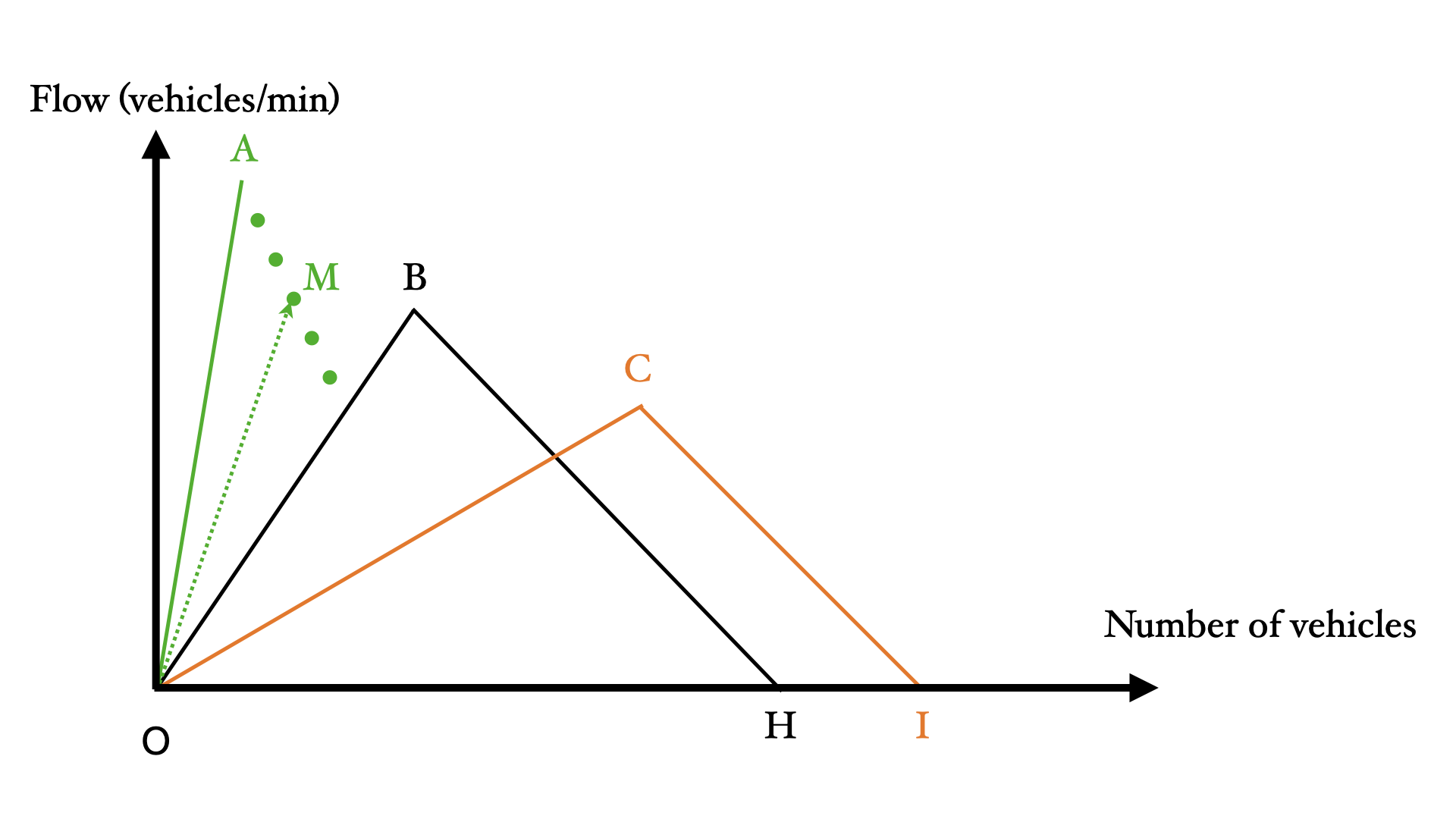}
  \caption{Traffic state during EMV pre-emption.}
  \label{fig_one_lane}
\end{figure}

During the EMV pre-emption, the original traffic flow relationship pictured in black are diverted into two parts, representing two lanes respectively. The green line represents the traffic conditions of the pre-emption lane, i.e. the lane where the EMV is traveling, and the orange line represents the other lane. During pre-emption, part of the vehicles originally travelling in front of the EMV pull over onto the adjacent lane, resulting a significant decrease in the max capacities of the pre-emption lane. Meanwhile, because vehicles can park onto the curbs, orange line depicts a larger maximum capacity $I$ than normal.

Regarding the max flow, the adjacent lane has a smaller max flow as the vehicles on this lane are required to slow down when EMV is on duty.
The pre-emption lane obtains a max capacity when there is one vehicle on the lane, i.e. the EMV itself. Under this circumstance, the flow rate is equivalent as the free flow speed of the EMV. However, when there are vehicles remaining in front of the EMV, the travel speed of the EMV might be slowed down, but still higher than the free flow speeds of the non-EMVs. Furthermore, the travel speed of EMV has a discrete value corresponding to the number of non-EMVs blocking. For example, pre-emption lane has a traffic state represented by $M$, and the travel speed of the EMV is obtained as the slope of $M$. We use this simple model, especially the green plot to estimate the intra-link travel time of EMV in a link as a function of number of vehicles in that link. 
We calibrate this model with the SUMO environment. Intuitively, the travel speed of the EMV is affected more by the number of vehicles on the pre-emption lane than their positions. The reason behind is since the ETA is frequently updated every $\Delta T$ seconds ($\Delta T$ in our experiment), and the estimated ETA would eventually converge. Imagine there are vehicles very far away from the EMV and about to leave the intersections at $t$, their presence would not slow down the approaching EMV. Therefore, when they have left the intersection at $t+1$, we have an updated number of non-EMVs count and updated travel speed estimation.
\section{Implementation Details}
Although MDP step length can be arbitrarily small enough for optimality, traffic signal phases should maintained a minimum amount of time so that vehicles and pedestrians can safely cross the intersections. To avoid rapid switching between the phases, we set our MDP time step length to be 5 seconds.


\subsection{Implementation details for synthetic $\text{grid}_{5 \times 5}$}
\begin{itemize}
    \item dimension of $s^t_{\mathcal{V}_i}$: $5 \times (8+8+4+2)=110$
    \item dimension of $\Tilde{s}^t_{\mathcal{V}_i}$: $5 \times (8+8+4+2)=110$
    \item dimension of $\pi^{t-1}_{\mathcal{N}_i}$: $4 \times 8 = 32$
    \item Policy network $\pi_{\theta_i}(a_i^t|s^t_{\mathcal{V}_i}, \pi^{t-1}_{\mathcal{N}_i})$: 
    \texttt{concat[}$110 \xrightarrow[]{\textrm{FC}} 128$ReLu, $32 \xrightarrow[]{\textrm{FC}} 64$ReLu\texttt{]} $ \xrightarrow[]{} 64$LSTM $ \xrightarrow[]{\textrm{FC}}8$Softmax
    \item Value network $V_{\phi_i}(\Tilde{s}^t_{\mathcal{V}_i}, \pi^{t-1}_{\mathcal{N}_i})$: \texttt{concat[}$110 \xrightarrow[]{\textrm{FC}} 128$ReLu, $32 \xrightarrow[]{\textrm{FC}} 64$ReLu\texttt{]} $ \xrightarrow[]{} 64$LSTM $ \xrightarrow[]{\textrm{FC}}1$Linear
    \item Each link is $200m$. The free flow speed of the EMV is $12m/s$ and the free flow speed for non-EMVs is $6m/s$.
    \item Temporal discount factor $\gamma$ is $0.99$ and spatial discount factor $\alpha$ is $0.90$.
    \item Initial learning rates $\eta_\phi$ and $\eta_\theta$ are both 1e-3 and they decay linearly. Adam optimizer is used.
    \item MDP step length $\Delta t = 5s$ and for secondary pre-emption reward weight $\beta$ is $0.5$.
    \item Regularization coefficient is $0.01$.
\end{itemize}
\subsection{Implementation details for $\text{Manhattan}_{16 \times 3}$}
The implementation is similar to the synthetic network implementation, with the following differences:
\begin{itemize}
    \item Initial learning rates $\eta_\phi$ and $\eta_\theta$ are both 5e-4.
    \item Since the avenues and streets are both one-directional, the number of actions of each agent are adjusted accordingly. 
\end{itemize}
\section{Ablation Study}
\subsection{Ablation study on pressure and agent types}
We propose three types of agents and design their rewards (Eqn.~1) based on our improved pressure definition and heuristics. 
In order to see how our improved pressure definition and proposed special agents influence the results, we (1) replace our pressure definition by that defined in PressLight, (2) replace secondary pre-emption agents with normal agents and (3) replace primary pre-emption agents with normal agents. 
\begin{table}[h!]
\centering
\fontsize{9.0pt}{10.0pt} \selectfont
\begin{tabular}{@{}cccc|c@{}}
\toprule[1pt]
Ablations                        & (1) & (2) & (3) & EMVLight\\ \midrule
$T_{\text{EMV}}$ [s]         & \textbf{197}       & 289                     & 320     & 199       \\
$T_{\text{avg}}$ [s]  & 361.05    & 347.13                  & 359.62   & \textbf{322.40}      \\ \bottomrule[1pt]
\end{tabular}
\caption{Ablation study on pressure and agent types. Experiments are conducted on the Config 1 synthetic $\text{grid}_{5 \times 5}$.}
\label{tab_ablation_reward}
\end{table}

Table \ref{tab_ablation_reward} shows the results of these ablations: (1) PressLight-style pressure (see Appendix) yields a slightly smaller EMV travel time but significantly increases the average travel time; (2) Without secondary pre-emption agents, EMV travel time increases by 45\% since almost no ``link reservation" happened; (3) Without primary pre-emption agents, EMV travel time increases significantly, which shows the importance of pre-emption.

\subsection{Ablation study on fingerprint}
\begin{figure}[ht]
    \centering
    \includegraphics[width=0.75\linewidth]{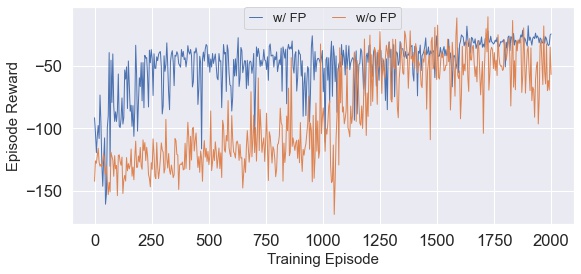}
  \caption{Reward convergence with and without fingerprint. Experiments are conducted on Config 1 synthetic $\text{grid}_{5 \times 5}$.}
  \label{fig_FP_comparison}
\end{figure}

In multi-agent RL, fingerprint has been shown to stabilize training and enable faster convergence.
In order to see how fingerprint affects training in EMVLight, we remove the fingerprint design, i.e., policy and value networks are changed from $\pi_{\theta_i}(a_i^t|s^t_{\mathcal{V}_i}, \pi^{t-1}_{\mathcal{N}_i})$ and  $V_{\phi_i}(\Tilde{s}^t_{\mathcal{V}_i}, \pi^{t-1}_{\mathcal{N}_i})$ to $\pi_{\theta_i}(a_i^t|s^t_{\mathcal{V}_i})$ and  $V_{\phi_i}(\Tilde{s}^t_{\mathcal{V}_i})$, respectively. 
Fig.~\ref{fig_FP_comparison} shows the influence of fingerprint on training. With fingerprint, the reward converges faster and suffers from less fluctuation, confirming the effectiveness of fingerprint. 

\end{document}